\documentclass{article}
\usepackage{ijcai15}
\usepackage{subcaption}
\usepackage{graphicx} 
\usepackage{algorithm}
\usepackage{algorithmic}
\usepackage{amsmath}
\usepackage{amsthm}
\usepackage{amssymb}
\usepackage{hhline}
\usepackage{enumitem}
\usepackage{times}
\usepackage{xcolor}

\newcommand{\RR}{\mathbb{R}}

\theoremstyle{definition}

\setlist{nolistsep}

\title{Self-Adaptive Hierarchical Sentence Model}
\author{Han Zhao$^\dagger$ \quad Zhengdong Lu$^\S$ \and Pascal Poupart$^\dagger$ \\
$^\dagger$David R. Cheriton School of Computer Science, University of Waterloo, ON, Canada\\
$^\S$Noah's Ark Lab, Huawei Technologies, Shatin, HongKong \\
$^\dagger$\{han.zhao, ppoupart\}@uwaterloo.ca, $^\S$lu.zhengdong@huawei.com
}

\begin{document}
\maketitle

\begin{abstract}
The ability to accurately model a sentence at varying stages (e.g., word-phrase-sentence) plays a central role in natural language processing. As an effort towards this goal we propose a self-adaptive hierarchical sentence model (AdaSent). AdaSent effectively forms a hierarchy of representations from words to phrases and then to sentences through recursive gated local composition of adjacent segments. We design a competitive mechanism (through gating networks) to allow the representations of the same sentence to be engaged in a particular learning task (e.g., classification), therefore effectively mitigating the gradient vanishing problem persistent in other recursive models.  Both qualitative and quantitative analysis shows that AdaSent can automatically form and select the representations suitable for the task at hand during training, yielding superior classification performance over competitor models on 5 benchmark data sets.
\end{abstract}

\section{Introduction}
The goal of sentence modeling is to represent the meaning of a sentence so that it can be used as input for other tasks. Previously, this task was often cast as semantic parsing, which aims to find a logical form that can describe the sentence. With recent advances in distributed representations and deep neural networks, it is now common practice to find a vectorial representation of sentences, which turns out to be quite effective for tasks of classification~\cite{kim2014convolutional}, machine translation~\cite{cho2014properties,bahdanau2014neural}, and semantic matching~\cite{hu2014convolutional}.

Perhaps the simplest method in this direction is the continuous Bag-of-Words (cBoW), where the representations of sentences are obtained by global pooling (e.g, average-pooling or max-pooling) over their word-vectors. The word-vectors, also known as word-embedding, can be determined in either supervised or unsupervised fashion. cBoW, although effective at capturing the topics of sentences, does not consider the sequential nature of words, and therefore has difficulty capturing the structure of sentences. There has been a surge of sentence models with the order of words incorporated, mostly based on neural networks of various forms, including recursive neural networks~\cite{socher2010learning,socher2012semantic,socher2013recursive}, recurrent neural network~\cite{irsoy2014deep,lai2015recurrent}, and convolution neural network~\cite{KalchbrennerACL2014,kim2014convolutional}. These works apply levels of non-linear transformations to model interactions between words and the structure of these interactions can also be learned on the fly through gated networks~\cite{cho2014properties}.  However these models output a fixed length continuous vector that does not retain intermediate information obtained during the composition process, which may be valuable depending on the task at hand.

\begin{figure}[htb]
\centering
	\includegraphics[width=\linewidth]{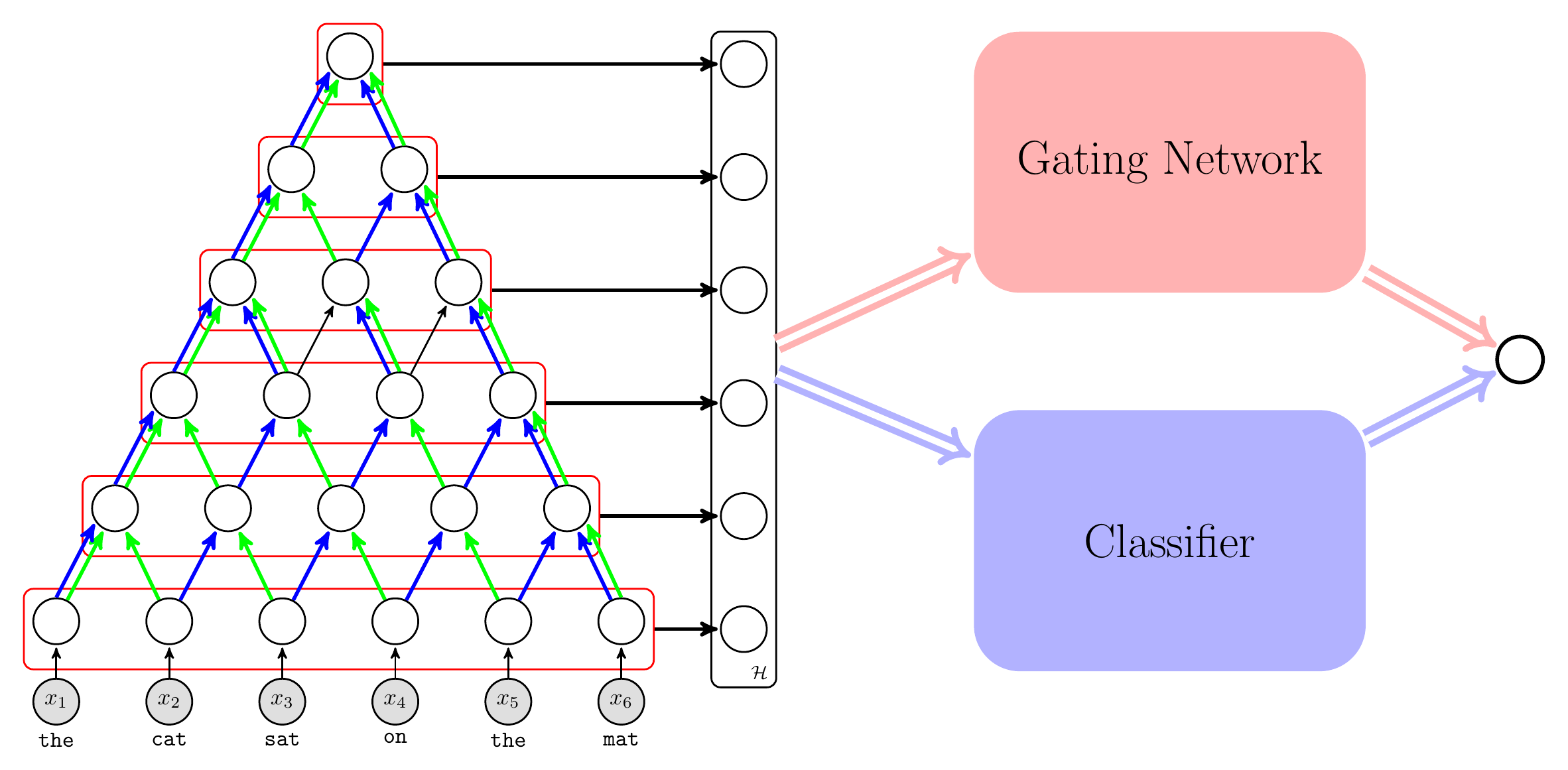}
\caption{The overall diagram of AdaSent (better viewed in color).  Flows with green and blue colors act as special cases for recurrent neural networks and recursive neural networks respectively (see more details in Sec. 3.2). Each level of the pyramid is pooled and the whole pyramid reduces into a hierarchy $\mathcal{H}$, which is then fed to a gating network and a classifier to form an ensemble.}
\label{fig:model}
\end{figure}

In this paper, we propose a self-adaptive hierarchical sentence model (AdaSent). Instead of maintaining a fixed-length continuous vectorial representation, our model forms a multi-scale hierarchical representation.  AdaSent is inspired from the gated recursive convolutional neural network (grConv)~\cite{cho2014properties} in the sense that the information flow forms a pyramid with a directed acyclic graph structure where local words are gradually composed to form intermediate representations of phrases.  Unlike cBoW, recurrent and recursive neural networks with fixed structures, the gated nature of AdaSent allows the information flow to vary with each task (i.e., no need for a pre-defined parse tree).  Unlike grConv, which outputs a fixed-length representation of the sentence at the top of the pyramid, AdaSent uses the intermediate representations at each level of the pyramid to form a multiscale summarization.  A convex combination of the representations at each level is used to adaptively give more weight to some levels depending on the sentence and the task.  Fig.~\ref{fig:model} illustrates the architecture of AdaSent and compares it to cBoW, recurrent neural networks and recursive neural networks.

Our contributions can be summarized as follows. First, we propose a novel architecture for short sequence modeling which explores a new direction to use a hierarchical multiscale representation rather than a flat, fixed-length representation. Second, we qualitatively show that our model is able to automatically learn the representation which is suitable for the task at hand through proper training. Third, we conduct extensive empirical studies on 5 benchmark data sets to quantitatively show the superiority of our model over previous approaches.

\section{Background}
\label{sec:background}
Let $\mathbf{x}_{1:T}$ denote the input sequence with length $T$. Each token $x_t \in \mathbf{x}_{1:T}$ is a $V$ dimensional one-hot binary vector to encode the $i$th word, where $V$ is the size of the vocabulary. We use $U\in\RR^{d\times V}$ to denote the word embedding matrix, in which the $j$th column is the $d$-dimensional distributed representation of the $j$th word in the vocabulary. Hence the word vectors for the sequence $\mathbf{x}_{1:T}$ is obtained by $\mathbf{h}^0_{1:T} = U\mathbf{x}_{1:T}$.

In the cBoW sentence model, the representation $\bar{h}$ for $\mathbf{x}_{1:T}$ is obtained by global pooling, either average pooling (Eq.~\ref{equ:average}) or max pooling (Eq.~\ref{equ:max}), over all the word vectors:
\begin{equation}
\label{equ:average}
\bar{h} = \frac{1}{T}\sum_{t=1}^T h_t^0 = \frac{U}{T}\sum_{t=1}^T x_t
\end{equation}
\begin{equation}
\label{equ:max}
\bar{h}_j = \max_{t\in 1:T}{h_t^0}_j, \quad j=1,\ldots,d
\end{equation}
It is clear that cBoW is insensitive to the ordering of words and also the length of a sentence, hence it is likely for two different sentences with different semantic meanings to be embedded into the same vector representation.

Recurrent neural networks~\cite{elman1990finding} are a class of neural networks where recurrent connections between input units and hidden units are formed through time. The sequential nature of recurrent neural networks makes them applicable to various sequential generation tasks, e.g., language modeling~\cite{mikolov2010recurrent} and machine translation~\cite{bahdanau2014neural,cho2014properties}.

\begin{figure}[htb]
\centering
	\begin{subfigure}[b]{0.4\textwidth}
		\includegraphics[width=\textwidth]{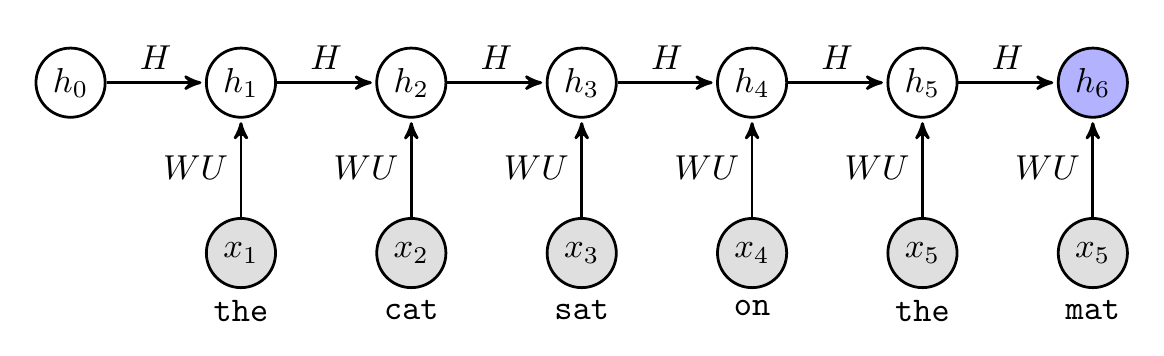}
		\caption{Composition process in a recurrent neural network.}
		\label{fig:recurrent}
	\end{subfigure}
	~
	\begin{subfigure}[b]{0.4\textwidth}
		\includegraphics[width=\textwidth]{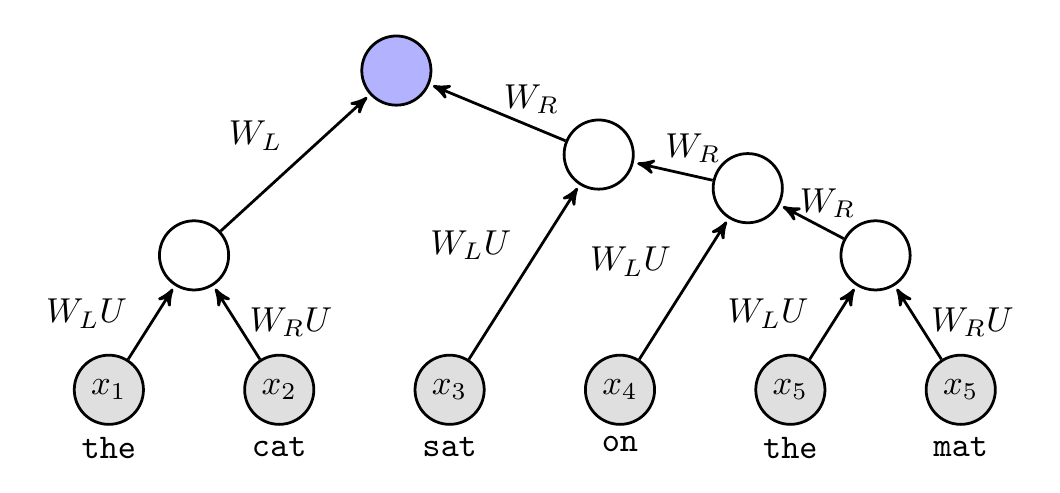}
		\caption{Composition process in a recursive neural network.}
		\label{fig:recursive}
	\end{subfigure}
\caption{Composition dynamics in recurrent and recursive neural networks. The one-hot binary encoding of word sequences is first multiplied by the word embedding matrix $U$ to obtain the word vectors before entering the network.}
\label{fig:nn}
\end{figure}

Given a sequence of word vectors $\mathbf{h}_{1:T}^0$, the hidden layer vector $h_t$ at time step $t$ is computed from a non-linear transformation of the current input vector $h_t^0$ and the hidden vector at the previous time step $h_{t-1}$. Let $W$ be the input-hidden connection matrix, $H$ be the recurrent hidden-hidden connection matrix and $b$ be the bias vector. Let $f(\cdot)$ be the component-wise non-linear transformation function. The dynamics of recurrent neural networks can be described by the following equations:
\begin{equation}
\left\lbrace
\begin{array}{r l}
h_0 & = 0 \\
h_t & = f(Wh_{t}^0 + Hh_{t-1} + b)
\end{array}
\right.
\end{equation}
The sentence representation $\bar{h}$ is then the hidden vector obtained at the last time step, $h_{T}$, which summarizes all the past words. The composition dynamics in recurrent neural networks can be described by a chain as in Fig.~\ref{fig:recurrent}.

Recursive neural networks build on the idea of composing along a pre-defined binary parsing tree. The leaves of the parsing tree correspond to words, which are initialized by their word vectors. Non-linear transformations are recursively applied bottom-up to generate the hidden representation of a parent node given the hidden representations of its two children. The composition dynamics in a recursive neural network can be described as $h = f(W_Lh_l + W_R h_r + b)$, where $h$ is the hidden representation for a parent node in the parsing tree and $h_l$, $h_r$ are the hidden representations for the left and right child of the parent node, respectively. $W_L$, $W_R$ are left and right recursive connection matrices. Like in recurrent neural networks, all the parameters in recursive neural networks are shared globally. The representation for the whole sentence is then the hidden vector obtained at the root of the binary parsing tree. An example is shown in Fig.~\ref{fig:recursive}.

Although the composition process is nonlinear in recursive neural network, it is pre-defined by a given binary parsing tree. Gated recursive convolutional neural network (grConv)~\cite{cho2014properties} extends recursive neural network through a gating mechanism to allow it to learn the structure of recursive composition on the fly. If we consider the composition structure in a recurrent neural network as a linear chain and the composition structure in a recursive neural network as a binary tree, then the composition structure in a grConv can be described as a pyramid, where word representations are locally combined until we reach the top of the pyramid, which gives us the global representation of a whole sentence. We refer interested readers to~\cite{cho2014properties} for more details about grConv.

\section{Self-Adaptive Hierarchical Sentence Model}
\label{sec:adasent}
AdaSent is inspired and built based on grConv. AdaSent differs from grConv and other neural sentence models that try to obtain a fixed-length vector representation by forming a hierarchy of abstractions of the input sentence and by feeding the hierarchy as a multi-scale summarization into the following classifier, combined with a gating network to decide the weight of each level in the final consensus, as illustrated in Fig.~\ref{fig:model}. 

\subsection{Structure}
\begin{figure}[htb]
\centering
	\includegraphics[scale=0.25]{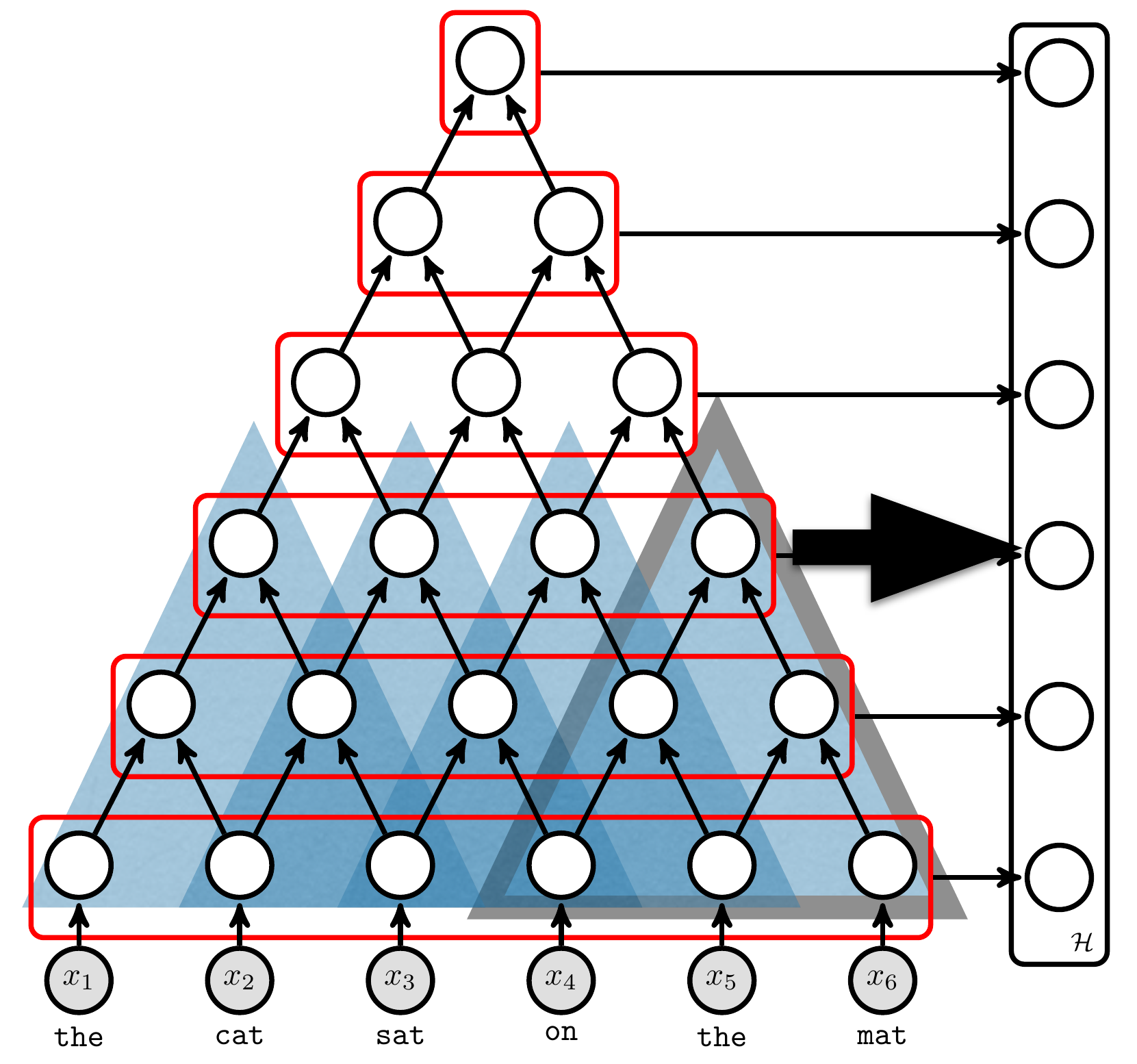}
\caption{Composition dynamics in AdaSent. The $j$th unit on the $t$th level is an intermediate hidden representation of the phrase $\mathbf{x}_{j:j+t-1}$ in the original sentence. All the units on the $t$th level are then pooled to obtain the $t$th level representation in the hierarchy $\mathcal{H}$.}
\label{fig:illustration}
\end{figure}

The structure of AdaSent is a directed acyclic graph as shown in Fig.~\ref{fig:illustration}. For an input sequence of length $T$, AdaSent is a pyramid of $T$ levels. Let the bottom level be the first level and the top level be the $T$th level. Define the \emph{scope} of each unit in the first layer to be the corresponding word, i.e., $\text{scope}(h^1_j) = \{x_j\}, \forall j\in 1:T$ and for any $t\geq 2$, define $\text{scope}(h^t_j) = \text{scope}(h^{t-1}_j)\cup\text{scope}(h^{t-1}_{j+1})$. Then the $t$th level in AdaSent contains a layer of $T-t+1$ units where each unit has a scope of size $t$. More specifically, the scope of $h^t_j$ is $\{\mathbf{x}_{j:j+t-1}\}$. Intuitively, for the sub-pyramid rooted at $h^t_j$, we can interpret $h^t_j$ as a top level summarization of the phrase $\mathbf{x}_{j:j+t-1}$ in the original sentence. For example, $h^3_4$ in Fig.~\ref{fig:illustration} can be viewed as a summarization of the phrase \texttt{on the mat}.

In general, units at the $t$th level are intermediate hidden representations of all the consecutive phrases of length $t$ in the original sentence (see the scopes of units at the 3rd level in Fig.~\ref{fig:illustration} for an example). There are two extreme cases in AdaSent: the first level contains word vectors and the top level is a global summarization of the whole sentence.

Before the pre-trained word vectors enter into the first level of the pyramid, we apply a linear transformation to map word vectors from $\RR^d$ to $\RR^D$ with $D\geq d$. That way we can allow phrases and sentences to be in a space of higher dimension than words for their richer structures.
More specifically, the hidden representation $\mathbf{h}^1_{1:T}$ at the first level of the pyramid is
\begin{equation}
\mathbf{h}^1_{1:T} = U'\mathbf{h}^0_{1:T} = U'U\mathbf{x}_{1:T}
\end{equation}
where $U'\in\RR^{D\times d}$ is the linear transformation matrix in AdaSent and $U\in\RR^{d\times V}$ is the word-embedding matrix trained with a large unlabeled corpus. Equivalently, one can view $\widetilde{U}\triangleq U'U\in\RR^{D\times V}$ as a new word-embedding matrix tailored for AdaSent. This factorization of the word-embedding matrix also helps to reduce the effective number of parameters in our model when $d\ll D$.

\subsection{Local Composition and Level Pooling}
The recursive local composition in the pyramid works in the following way
\begin{equation}
h_j^t = \omega_l h_{j}^{t-1} + \omega_r h_{j+1}^{t-1} + \omega_c \tilde{h}_j^t
\label{equ:gate}
\end{equation}
\begin{equation}
\tilde{h}_j^t = f(W_L h_j^{t-1} + W_R h_{j+1}^{t-1} + b_W)
\label{equ:comp}
\end{equation}
where $j$ ranges from 1 to $T-t+1$ and $t$ ranges from $2$ to $T$. $W_L, W_R\in\RR^{D\times D}$ are the hidden-hidden combination matrices, dubbed recurrent matrices, and $b_W\in\RR^D$ is a bias vector. $\omega_l$, $\omega_r$ and $\omega_c$ are the gating coefficients which satisfy $\omega_l, \omega_r, \omega_c\geq 0$ and $\omega_l + \omega_r + \omega_c = 1$. Eq.~\ref{equ:comp} provides a way to compose the hidden representation of a phrase of length $t$ from the hidden representation of its left $t-1$ prefix and its right $t-1$ suffix. The composition in Eq.~\ref{equ:comp} includes a non-linear transformation, which allows a flexible hidden representation to be formed. The fundamental assumption behind the structure of AdaSent is then encoded in Eq.~\ref{equ:gate}: the semantic meaning of a phrase of length $t$ is a convex combination of the semantic meanings of its $t-1$ prefix, $t-1$ suffix and the composition of these two. For example, we expect the meaning of the phrase \texttt{the cat} to be expressed by the word \texttt{cat} since \texttt{the} is only a definite article, which does not have a direct meaning. On the other hand, we also hope the meaning of the phrase \texttt{not happy} to consider both the functionality of \texttt{not} and also the meaning of \texttt{happy}. We design the local composition in AdaSent to make it flexible enough to catch the above variations in language while letting the gating mechanism (the way to obtain $\omega_l,\omega_r$ and $\omega_c$) adaptively decide the most appropriate composition from the current context.

Technically, when computing $h_j^t$, $\omega_l, \omega_c$ and $\omega_r$ are parametrized functions of $h_j^{t-1}$ and $h_{j+1}^{t-1}$ such that they can decide whether to compose these two children by a non-linear transformation or simply to forward the children's representations for future composition. For the purpose of illustration, we use the \texttt{softmax} function to implement the gating mechanism during the local composition in Eq.~\ref{equ:local}. But note that we are not limited to a specific choice of gating mechanism. One can adopt more complex systems, e.g., MLP, to implement the local gating mechanism as long as the output of the system is a multinomial distribution over 3 categories.
\begin{equation}
\label{equ:local}
\begin{pmatrix}
w_l \\ w_r \\ w_c
\end{pmatrix} = \text{\texttt{softmax}}(G_L h_j^{t-1} + G_R h_{j+1}^{t-1} + b_G)
\end{equation}
$G_L, G_R\in\RR^{3\times D}$ and $b_G\in\RR^{3}$ are shared globally inside the pyramid. The \texttt{softmax} function over a vector is defined as:
\begin{equation} 
\text{\texttt{softmax}}(\mathbf{v}) = \frac{1}{\sum_{i=1}^l\exp(v_i)}
\begin{pmatrix} 
\exp(v_1) \\ \vdots \\ \exp(v_l)
\end{pmatrix}, \quad \mathbf{v}\in\RR^l
\end{equation}
Local compositions are recursively applied until we reach the top of the pyramid.

\begin{figure}[htb]
\centering
	\includegraphics[scale=0.4]{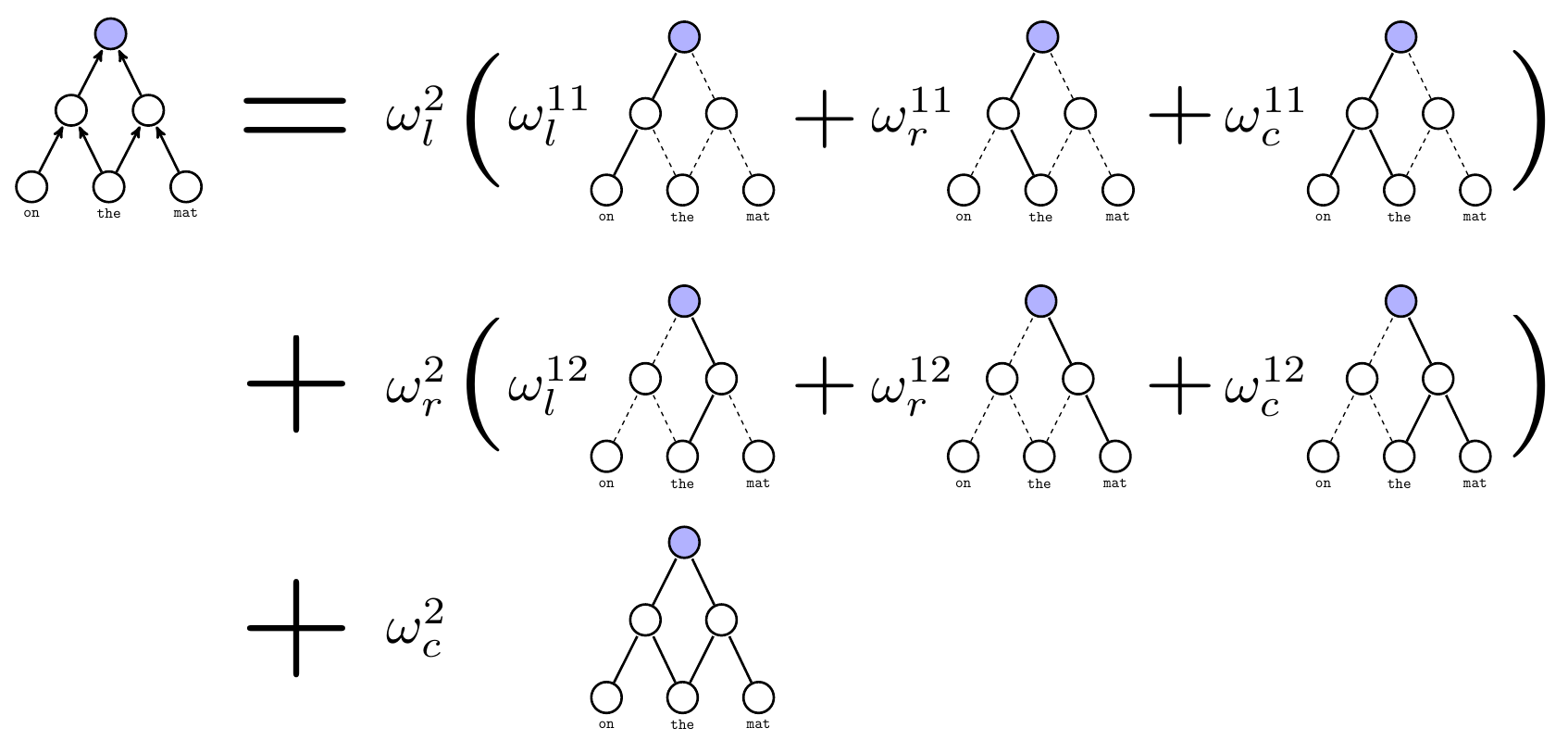}
\caption{The hidden vector obtained at the top can be decomposed into a convex combination of all possible hidden vectors composed along the corresponding sub-structures.}
\label{fig:composition}
\end{figure}

It is worth noting that the recursive local composition in AdaSent implicitly forms a weighted model average such that each unit at layer $t$ corresponds to a convex combination of all possible sub-structures along which the composition process is applied over the phrase of length $t$. This implicit weighted model averaging makes AdaSent more robust to local noises and deteriorations than recurrent nets and recursive nets where the composition structure is unique and rigid. Fig.~\ref{fig:composition} shows an example when $t=3$. 

Once the pyramid has been built, we apply a pooling operation, either average pooling or max pooling, to the $t$th level, $t\in 1:T$, of the pyramid to obtain a summarization of all consecutive phrases of length $t$ in the original sentence, denoted by $\bar{h}^t$ (see an example illustrated in Fig.~\ref{fig:illustration} for the global level pooling applied to the 3rd level in the pyramid).
It is straightforward to verify that $\bar{h}^1$ corresponds to the representation returned by applying cBoW to the whole sentence. $[(\bar{h}^1)^T, \cdots, (\bar{h}^T)^T]^T$ then forms the hierarchy in which lower level summarization in the hierarchy pays more attention to local words or short phrases while higher level summarization focuses more on the global interaction of different parts in the sentence.

\subsection{Gating Network}
Suppose we are interested in a classification problem, one can easily extend our approach to other problems of interests. Let $g(\cdot)$ be a discriminative classifier that takes $\bar{h}^t\in\RR^D$ as input and outputs the probabilities for different classes. Let $w(\cdot)$ be a gating network that takes $\bar{h}^t\in\RR^D, t=1, \ldots, T$ as input and outputs a belief score $0\leq\gamma_t\leq 1$. Intuitively, the belief score $\gamma_t$ depicts how confident the $t$th level summarization in the hierarchy is suitable to be used as a proper representation of the current input instance for the task at hand. We require $\gamma_t\geq 0, \forall t$ and $\sum_{t=1}^{T}\gamma_t = 1$.

Let $C$ denote the categorical random variable corresponding to the class label. The consensus of the whole system is reached by taking a mixture of decisions made by levels of summarizations from the hierarchy:
\begin{align}
p(C = c | \mathbf{x}_{1:T}) & = \sum_{t = 1}^{T}p(C = c | \mathcal{H}_{\mathbf{x}} = t)\cdot p(\mathcal{H}_{\mathbf{x}} = t | \mathbf{x}_{1:T}) \nonumber\\
&= \sum_{t=1}^Tg(\bar{h}^t)\cdot w(\bar{h}^t)
\end{align}
where each $g(\cdot)$ is the classifier and $w(\cdot)$ corresponds to the gating network in Fig.~\ref{fig:model}.

\subsection{Back Propagation through Structure}
We use back propagation through structure (BPTS)~\cite{goller1996learning} to compute the partial derivatives of the objective function with respect to the model parameters. Let $\mathcal{L}(\cdot)$ be our scalar objective function. The goal is to derive the partial derivative of $\mathcal{L}$ with respect to the model parameters in AdaSent, i.e., two recurrent matrices, $W_L$, $W_R$ and two local composition matrices $G_L, G_R$ (and their corresponding bias vectors):
\begin{equation} 
\frac{\partial \mathcal{L}}{\partial W_L}\hspace{-2pt} =\hspace{-2pt} \sum_{t = 1}^T\sum_{j=1}^{T-t+1}\frac{\partial \mathcal{L}}{\partial h^t_{j}}\frac{\partial h^t_j}{\partial W_L}, \frac{\partial\mathcal{L}}{\partial W_R} \hspace{-2pt} = \hspace{-2pt} \sum_{t = 1}^T\sum_{j=1}^{T-t+1}\frac{\partial \mathcal{L}}{\partial h^t_{j}}\frac{\partial h_j^t}{\partial W_R}
\end{equation}
The same analysis can be applied to compute $\frac{\partial \mathcal{L}}{\partial G_L}$ and $\frac{\partial \mathcal{L}}{\partial G_R}$. Taking into account the DAG structure of AdaSent, we can compute $\frac{\partial \mathcal{L}}{\partial h^t_j}$ recursively in the following way:
\begin{equation}
\frac{\partial \mathcal{L}}{\partial h^t_j} = \frac{\partial \mathcal{L}}{\partial h^{t+1}_j}\frac{\partial h^{t+1}_j}{\partial h^t_j} + \frac{\partial\mathcal{L}}{\partial h^{t+1}_{j-1}}\frac{\partial h^{t+1}_{j-1}}{\partial h^{t}_{j}}
\end{equation}
Now consider the left and right local BP formulations:
\begin{eqnarray}
\label{equ:leftbp}
\frac{\partial h^{t+1}_{j-1}}{\partial h^t_j} &=& \omega_r I + \omega_c\text{diag}(f')W_R\\
\frac{\partial h^{t+1}_j}{\partial h^t_j} &=& \omega_l I + \omega_c\text{diag}(f')W_L
\label{equ:rightbp} 
\end{eqnarray}
where $I$ is the identity matrix and $\text{diag}(f')$ is a diagonal matrix spanned by the vector $f'$, which is the derivative of $f(\cdot)$ with respect to its input. The identity matrix in Eq.~\ref{equ:leftbp} and Eq.~\ref{equ:rightbp} plays the same role as the linear unit recurrent connection in the memory block of LSTM~\cite{hochreiter1997long} to allow the \emph{constant error carousel} to effectively prevent the gradient vanishing problem that commonly exists in recurrent neural nets and recursive neural nets. Also, the local composition weights $\omega_l$, $\omega_r$ and $\omega_c$ in Eq.~\ref{equ:leftbp} and Eq.~\ref{equ:rightbp} have the same effect as the forgetting gate in LSTM~\cite{gers2000learning} by allowing more flexible credit assignments during the back propagation process.

\section{Experiments}
In this section, we study the empirical performance of AdaSent on 5 benchmark data sets for sentence and short phrase classification and then compare it to other competitor models. We also visualize the representation of the input sequence learned by AdaSent by projecting it in a 2 dimensional space using PCA to qualitatively study why AdaSent works for short sequence modeling. 
\label{sec:experiment}
\subsection{Experimental Setting}
Statistics about the data sets used in this paper are listed in Table~\ref{table:dataset}. We describe each data set in detail below:
\begin{enumerate}
	\item 	\textbf{MR}. Movie reviews~\cite{pang2005seeing}\footnote{https://www.cs.cornell.edu/people/pabo/movie-review-data/} data set where each instance is a sentence. The objective is to classify each review by its overall sentiment polarity, either positive or negative.
	\item 	\textbf{CR}. Annotated customer reviews of 14 products obtained from Amazon~\cite{hu2004mining}\footnote{http://www.cs.uic.edu/$\sim$liub/FBS/sentiment-analysis.html}. The task is to classify each customer review into positive and negative categories.
	\item 	\textbf{SUBJ}. Subjectivity data set where the goal is to classify each instance (snippet) as being subjective or objective~\cite{pang2004sentimental}.
	\item 	\textbf{MPQA}. Phrase level opinion polarity detection subtask of the MPQA data set~\cite{wiebe2005annotating}\footnote{http://mpqa.cs.pitt.edu/}.
	\item 	\textbf{TREC}. Question data set, in which the goal is to classify an instance (question) into 6 different types~\cite{li2002learning}\footnote{http://cogcomp.cs.illinois.edu/Data/QA/QC/}.
\end{enumerate}
\begin{table}[htb]
\centering
\begin{tabular}{c||c|c|c|c|c}\hline
\textbf{Data} & $N$ & \textbf{dist}(+,-) & $K$ & $|\mathbf{w}|$ & \textbf{test}\\\hline
MR & 10662 & (0.5, 0.5) & 2 & 18 & CV \\
CR & 3788 & (0.64, 0.36) & 2 & 17 & CV \\
SUBJ & 10000 & (0.5, 0.5) & 2 & 21 & CV \\
MPQA & 10099 & (0.31, 0.69) & 2 & 3 & CV \\
TREC & 5952 & \footnotesize{(0.1,0.2,0.2,0.1,0.2,0.2)} & 6 & 10 & 500 \\\hline
\end{tabular}
\caption{Statistics of the five data sets used in this paper. $N$ counts the number of instances and \textbf{dist} lists the class distribution in the data set. $K$ represents the number of target classes. $|\mathbf{w}|$ measures the average number of words in each instance. \textbf{test} is the size of the test set. For datasets which do not provide an explicit split of train/test, we use 10-fold cross-validation (CV) instead.}
\label{table:dataset}
\end{table}
We compare AdaSent with different methods listed below on the five data sets.
\begin{enumerate}
	\item 	\textbf{NB-SVM} and \textbf{MNB}. Naive Bayes SVM and Multinomial Naive Bayes with uni and bigram features~\cite{wang2012baselines}. 
	\item 	\textbf{RAE} and \textbf{MV-RecNN}. Recursive autoencoder~\cite{socher2011semi} and Matrix-vector recursive neural network~\cite{socher2012semantic}. In these two models, words are gradually composed into phrases and sentence along a binary parse tree.
	\item 	\textbf{CNN}~\cite{kim2014convolutional} and \textbf{DCNN}~\cite{KalchbrennerACL2014}. Convolutional neural network for sentence modeling. In DCNN, the author applies dynamic $k$-max pooling over time to generalize the original max pooling in traditional CNN.
	\item 	\textbf{P.V.}. Paragraph Vector~\cite{le2014distributed} is an unsupervised model to learn distributed representations of words and paragraphs. We use the public implementation of P.V.\footnote{https://github.com/mesnilgr/iclr15} and use logistic regression on top of the pre-trained paragraph vectors for prediction.
	\item	\textbf{cBoW}. Continuous Bag-of-Words model. As discussed above, we use average pooling or max pooling as the global pooling mechanism to compose a phrase/sentence vector from a set of word vectors.
	\item 	\textbf{RNN, BRNN}. Recurrent neural networks and bidirectional recurrent neural networks~\cite{schuster1997bidirectional}. 
For bidirectional recurrent neural networks, the reader is referred to \cite{lai2015recurrent} for more details.
	\item 	\textbf{GrConv}. Gated recursive convolutional neural network~\cite{cho2014properties} shares the pyramid structure with AdaSent and uses the top node in the pyramid as a fixed length vector representation of the whole sentence. 
\end{enumerate}

\subsection{Training}

The difficulty of training recurrent neural networks is largely due to the notorious gradient exploding and gradient vanishing problem~\cite{bengio1994learning,pascanu2013difficulty}. As analyzed and discussed before, the DAG structure combined with the local gating composition mechanism of AdaSent naturally help to avoid the gradient vanishing problem. However, the gradient exploding problem still exists as we observe in our experiments. In this section, we discuss our implementation details to mitigate the gradient exploding problem and we give some practical tricks to improve the performance in the experiments.
\subsubsection{Regularization of Recurrent Matrix}
The root of the gradient exploding problem in recurrent neural networks and other related models lies in the large spectral norm of the recurrent matrix as shown in Eq.~\ref{equ:leftbp} and Eq.~\ref{equ:rightbp}. Suppose the spectral norm of $W_L$ and $W_R$ $\gg 1$, then the recursive application of Eq.~\ref{equ:leftbp} and Eq.~\ref{equ:rightbp} in the back propagation process will cause the norm of the gradient vector to explode. To alleviate this problem, we propose to penalize the Frobenius norm of the recurrent matrix, which acts as a surrogate (upper bound) of the corresponding spectral norm, since 1) it is computationally expensive to compute the exact value of spectral norm and 2) it is hard to establish a direct connection between the spectral norm and the model parameters to incorporate it into our objective function. Let $\mathcal{L}(\cdot, \cdot)$ be our objective function to minimize. For example, when $\mathcal{L}$ is the negative log-likelihood in the classification setting, our optimization can be formulated as
\begin{equation}
\text{minimize}\quad \frac{1}{N}\sum_{i=1}^N\mathcal{L}(\mathbf{x}_i, y_i) + \lambda\left(||W_L||_F^2 + ||W_R||_F^2\right)
\label{equ:objective}
\end{equation}
where $\mathbf{x}_i$ is the training sequence and $y_i$ is the label. The value of the regularization coefficient $\lambda$ is problem dependent.  In our experiments, typical values of $\lambda$ range from $0.01$ to $5\times 10^{-5}$. For all our experiments, we use minibatch AdaGrad~\cite{duchi2011adaptive} with the norm-clipping technique~\cite{pascanu2013difficulty} to optimize the objective function in Eq.~\ref{equ:objective}.

\subsubsection{Implementation Details}
Throughout our experiments, we use a 50-dimensional word embedding trained using \texttt{word2vec}~\cite{mikolov2013distributed} on the Wikipedia corpus ($\sim$1B words). The vocabulary size is about 300,000. For all the tasks, we fine-tune the word embeddings during training to improve the performance~\cite{collobert2011natural}. We use the hyperbolic tangent function as the activation function in the composition process as the rectified linear units~\cite{nair2010rectified} are more prone to the gradient exploding problem in recurrent neural networks and its related variants. We use an MLP to implement the classifier on top of the hierarchy and use a softmax function to implement the gating network. We also tried using MLP to implement the gating network, but this does not improve the performance significantly.
\subsection{Experiment Results}
\begin{table}[htb]
\centering
\begin{tabular}{l||c|c|c|c|c}\hline
\textbf{Model} & MR & CR & SUBJ & MPQA & TREC \\\hline
NB-SVM & 79.4 & 81.8 & 93.2 & 86.3 & - \\
MNB & 79.0 & 80.0 & 93.6 & 86.3 & - \\
RAE &  77.7 & - & - & 86.4 & - \\
MV-RecNN & 79.0 & - & - & - & - \\
CNN & 81.5 & 85.0 & 93.4 & 89.6 & \textbf{93.6} \\
DCNN & - & - & - & - & 93.0 \\\hline
P.V. & $74.8$ & $78.1$ & $90.5$ & $74.2$ & $91.8$ \\\hline
cBoW & $77.2$ & $79.9$ & $91.3$ & $86.4$ & $87.3$ \\
RNN & $77.2$ & $82.3$ & $93.7$ & $90.1$ & $90.2$ \\
BRNN & $82.3$ & $82.6$ & $94.2$ & $90.3$ & $91.0$ \\
GrConv & $76.3$ & $81.3$ & $89.5$ & $84.5$ & $88.4$ \\\hline
AdaSent & $\mathbf{83.1}$ & $\mathbf{86.3}$ & $\textbf{95.5}$ & $\mathbf{93.3}$ & $92.4$\\\hline
\end{tabular}
\caption{Classification accuracy of AdaSent compared with other models. For NB-SVM, MNB, RAE, MV-RecNN, CNN and DCNN, we use the results reported in the corresponding paper. We use the public implementation of P.V. and we implement other methods.}
\label{table:result}
\end{table}

The classification accuracy of AdaSent compared with other models is shown in Table~\ref{table:result}. AdaSent consistently outperforms P.V., cBoW, RNN, BRNN and GrConv by a large margin while achieving comparable results to the state-of-the-art and using much fewer parameters: the number of parameters in our models range from 10K to 100K while in CNN the number of parameters is about 400K\footnote{The state-of-the-art accuracy on TREC is 95.0 achieved by~\cite{silva2011symbolic} using SVM with 60 hand-coded features.}. AdaSent outperforms all the other models on the MPQA data set, which consists of short phrases (the average length of each instance in MPQA is 3). We attribute the success of AdaSent on MPQA to its power in modeling short phrases since long range dependencies are hard to detect and represent.

Compared with BRNN, the level-wise global pooling in AdaSent helps to explicitly model phrases of different lengths while in BRNN the summarization process is more sensitive to a small range of nearby words.  Hence, AdaSent consistently outperforms BRNN on all data sets. Also, AdaSent significantly outperforms GrConv on all the data sets, which indicates that the variable length multi-scale representation is key to its success. As a comparison, GrConv does not perform well because it fails to keep the intermediate representations. More results on using GrConv as a fixed-length sequence encoder for machine translation and related tasks can be found in \cite{cho2014properties}. cBoW is quite effective on some tasks (e.g., SUBJ). We think this is due to the language regularities encoded in the word vectors and also the characteristics of the data itself. It is surprising that P.V. performs worse than other methods on the MPQA data set. This may be due to the fact that the average length of instances in MPQA is small, which limits the number of context windows when training P.V..
\begin{table}[htb]
\small
\centering
\begin{tabular}{l||c|c|c}\hline
\textbf{Model} & MR & CR & SUBJ \\\hline
P.V. & $71.11\pm 0.80$ & $71.22\pm 1.04$ & $90.22\pm 0.21$ \\\hline
cBoW & $72.74\pm 1.03$ & $71.86\pm 2.00$ & $90.58\pm 0.52$ \\
RNN & $74.39\pm 1.70$ & $73.81\pm 3.52$ & $89.97\pm 2.88$  \\
BRNN & $75.25\pm 1.33$ & $76.72\pm 2.78$ & $90.93\pm 1.00$ \\
GrConv & $71.64\pm 2.09$ & $71.52\pm 4.18$ & $86.53\pm 1.33$ \\\hline
AdaSent & $\mathbf{79.84\pm 1.26}$ & $\mathbf{83.61\pm 1.60}$ & $\mathbf{92.19\pm 1.19}$ \\\hline\hline
\textbf{Model} & MPQA & TREC \\\hline
P.V. & $67.93\pm  0.57$ & $86.30\pm 1.10$ & \\\hline
cBoW & $84.04\pm 1.20$ & $85.16\pm 1.76$ & \\
RNN & $84.52\pm 1.17$ & $84.24\pm 2.61$ & \\
BRNN & $85.36\pm 1.13$ & $86.28\pm 0.90$ & \\
GrConv & $82.00\pm 0.88$ & $82.04\pm 2.23$\\\hline
AdaSent & $\mathbf{90.42\pm 0.71}$ & $\mathbf{91.10\pm 1.04}$ & \\\hline
\end{tabular}
\caption{Model variance.}
\label{table:robustness}
\end{table}

We also report model variance of P.V., cBoW, RNN, BRNN, GrConv and AdaSent in Table~\ref{table:robustness} by running each of the models on every data set 10 times using different settings of hyper-parameters and random initializations. We report the mean classification accuracy and also the standard deviation of the 10 runs on each of the data set. Again, AdaSent consistently outperforms all the other competitor models on all the data sets.
\begin{figure}[htb]
\centering
	\includegraphics[width=\linewidth]{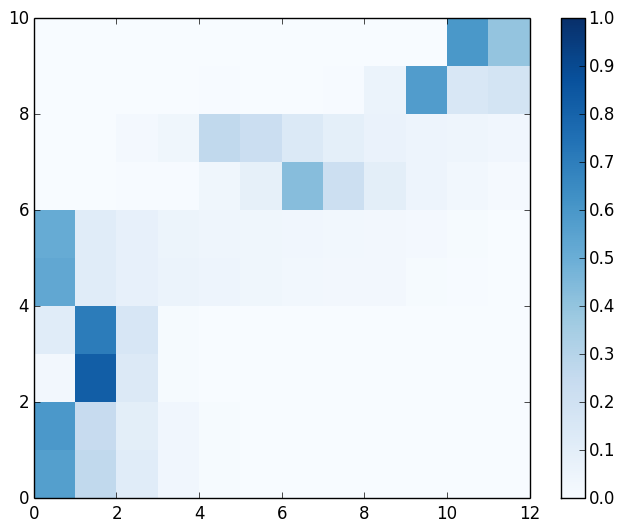}
\caption{Each row corresponds to the belief score of a sentence of length 12 sampled from one of the data sets.  From top to bottom, the 10 sentences are sampled from MR, CR, SUBJ, MPQA and TREC respectively.}
\label{fig:belief}
\end{figure}

\begin{figure}[htb]
\centering
	\includegraphics[width=\linewidth]{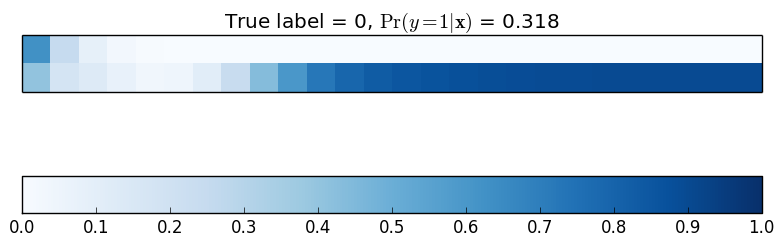}
\caption{Sentence: If the movie were all comedy it might work better but it has an ambition to say something about its subjects but not willingness.}
\label{fig:MR-exp}
\end{figure}

To study how the multi-scale hierarchy is combined by AdaSent in the final consensus, for each data set, we sample two sentences with a pre-specified length and compute their corresponding belief scores. We visualize the belief scores of 10 sentences by a matrix shown in Fig.~\ref{fig:belief}. As illustrated in Fig.~\ref{fig:belief}, the distribution of belief scores varies among different input sentences and also different data sets. The gating network is trained to adaptively select the most appropriate representation in the hierarchy by giving it the largest belief score. We also give a concrete example from MR to show both the predictions computed from each level and their corresponding belief scores given by the gating network in Fig.~\ref{fig:MR-exp}. The first row in Fig.~\ref{fig:MR-exp} shows the belief scores $\Pr(\mathcal{H}_{\mathbf{x}} = t|\mathbf{x}_{1:T}),\forall t$ and the second row shows the probability $\Pr(y=1|\mathcal{H}_{\mathbf{x}}=t),\forall t$ predicted from each level in the hierarchy. In this example, although the classifier predicts incorrectly for higher level representations, the gating network assigns the first level with the largest belief score, leading to a correct final consensus. The flexibility of multiscale representation combined with a gating network allows AdaSent to generalize GrConv in the sense that GrConv corresponds to the case where the belief score at the root node is 1.0.

\begin{figure}[htb]
\centering
\begin{subfigure}[b]{0.44\linewidth}
	\centering
	\includegraphics[width=\linewidth]{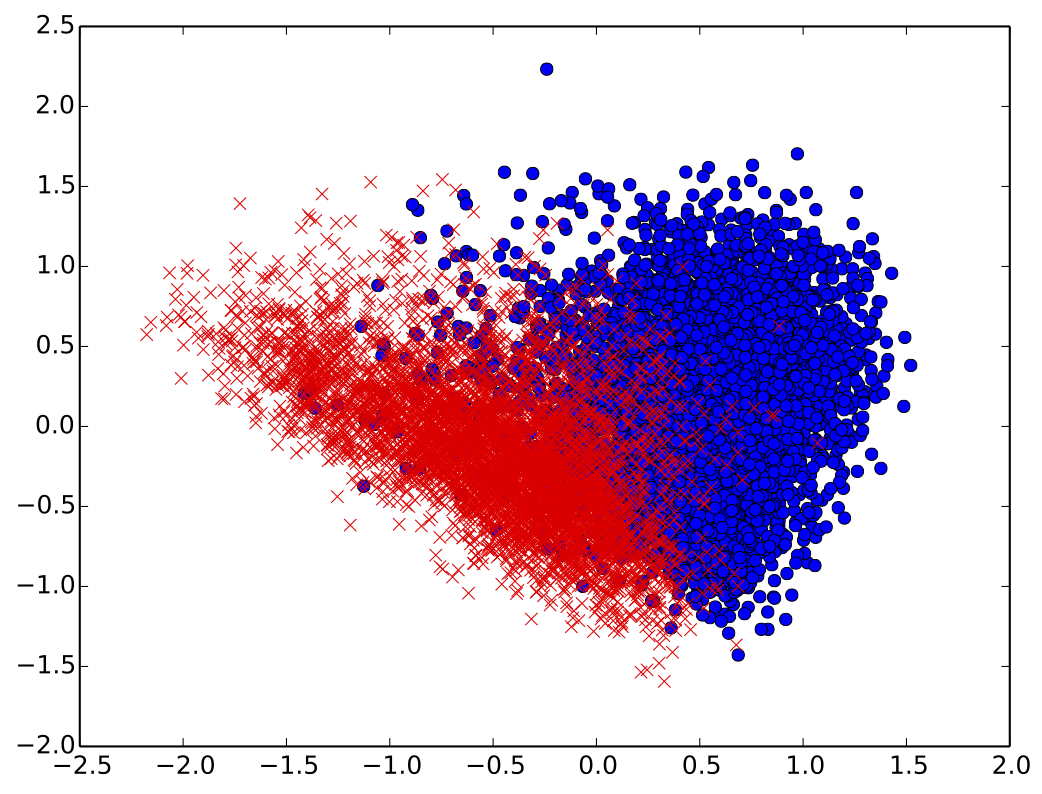}
\end{subfigure}
~
\begin{subfigure}[b]{0.44\linewidth}
	\centering
	\includegraphics[width=\linewidth]{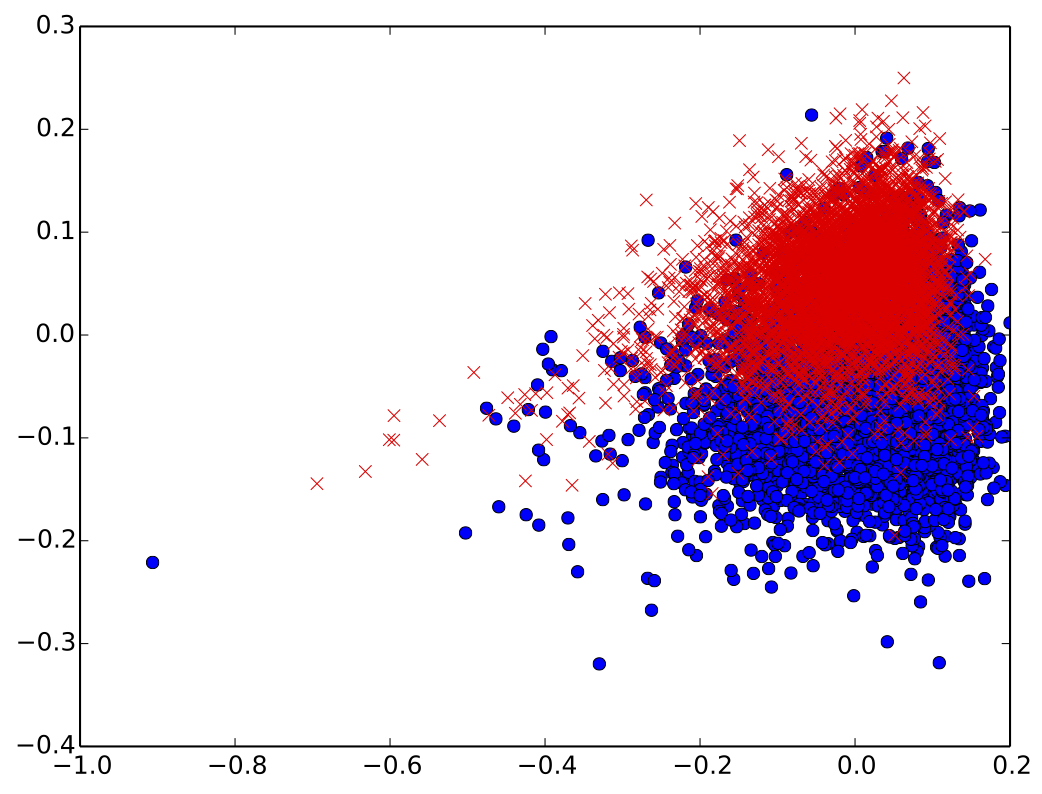}
\end{subfigure}
~
\begin{subfigure}[b]{0.44\linewidth}
	\centering
	\includegraphics[width=\linewidth]{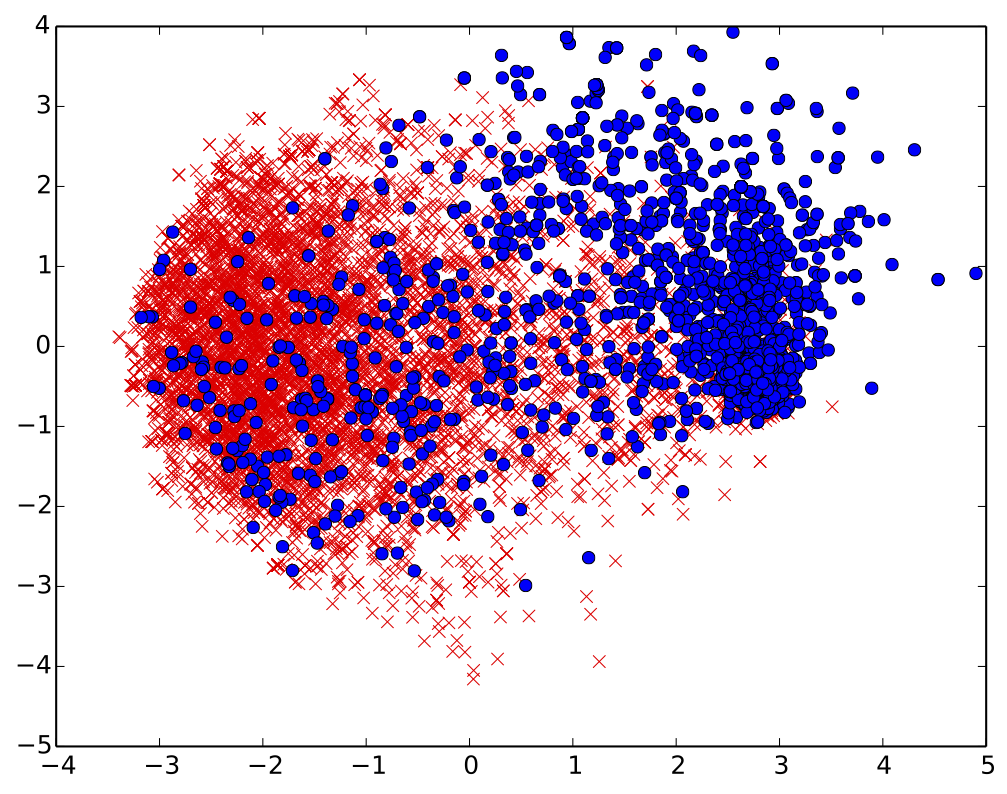}	
\end{subfigure}
~
\begin{subfigure}[b]{0.44\linewidth}
	\centering
	\includegraphics[width=\linewidth]{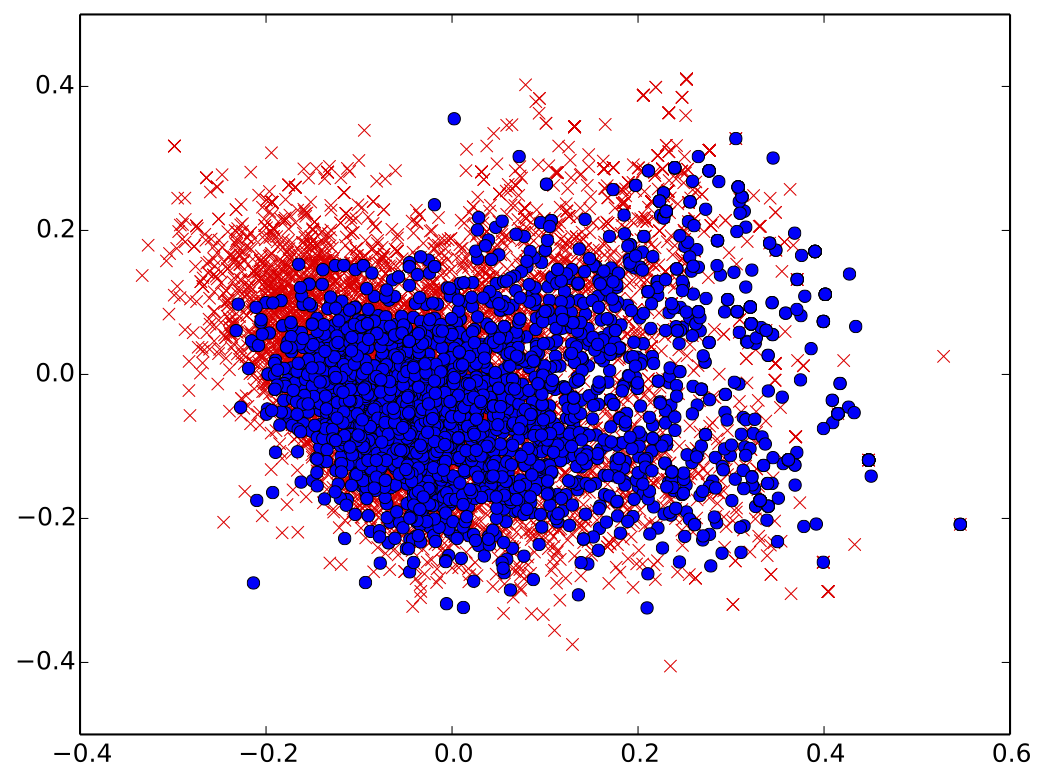}
\end{subfigure}
~
\begin{subfigure}[b]{0.44\linewidth}
	\centering
	\includegraphics[width=\linewidth]{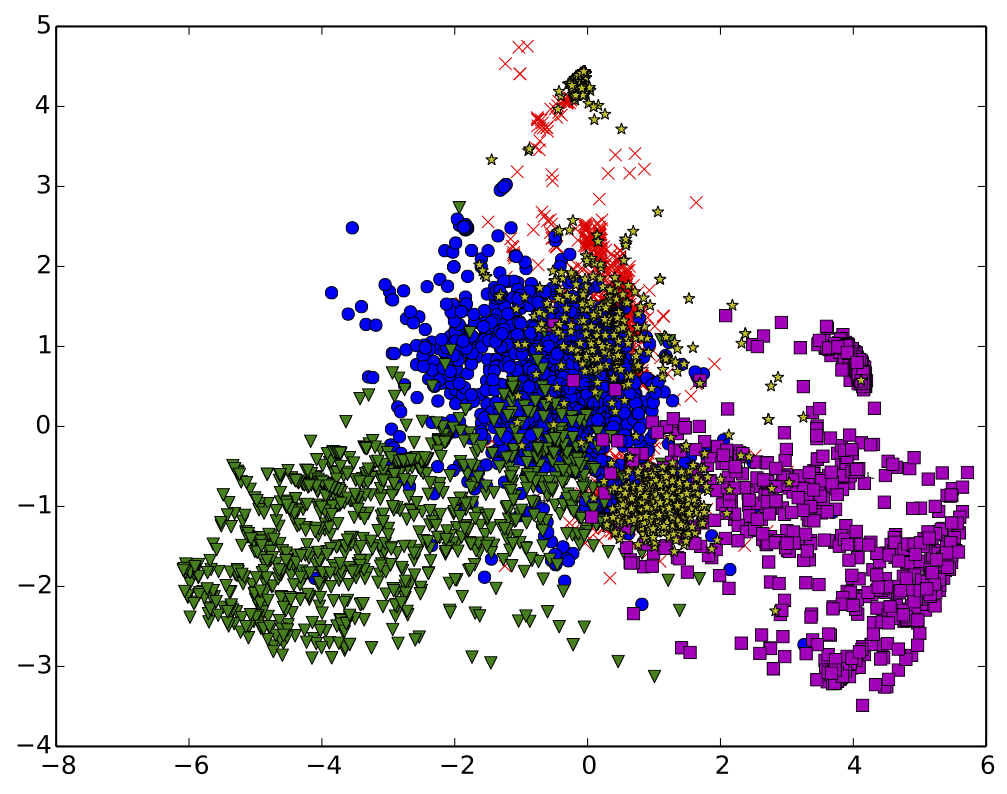}	
\end{subfigure}
~
\begin{subfigure}[b]{0.44\linewidth}
	\centering
	\includegraphics[width=\linewidth]{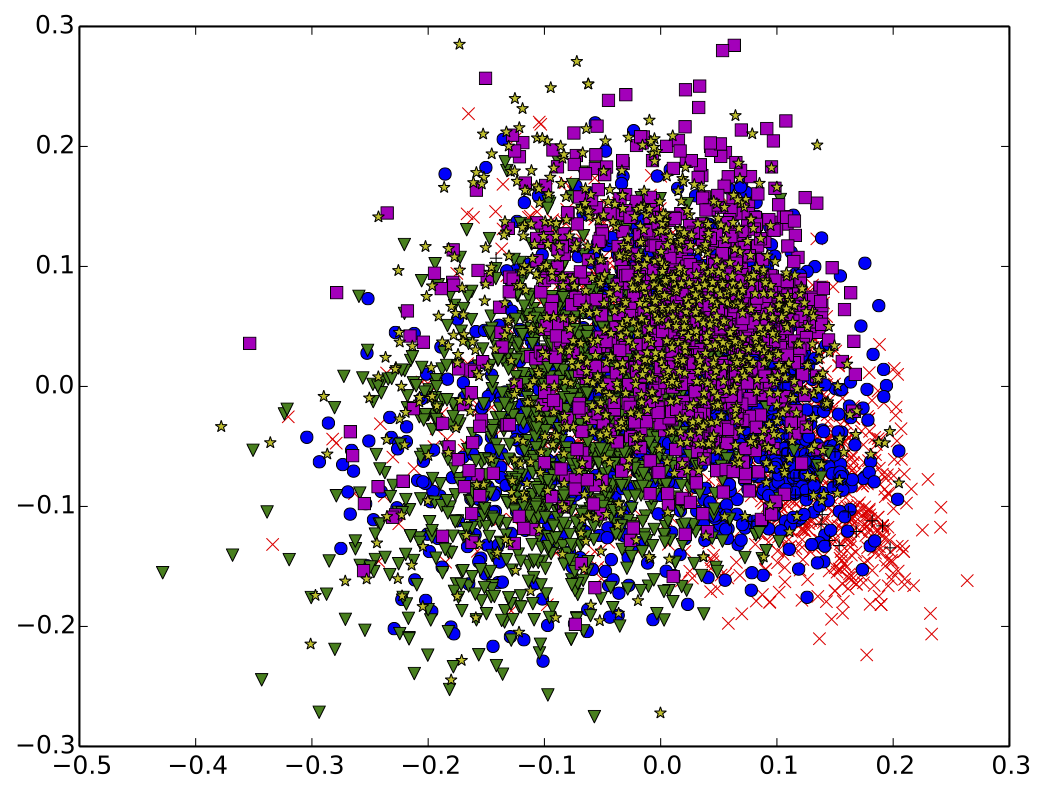}
\end{subfigure}
\caption{Different colors and patterns correspond to different objective classes. The first, second and third rows correspond to SUBJ, MPQA and TREC respectively and the left and right columns correspond to AdaSent and cBoW respectively.}
\label{fig:pca}
\end{figure}

To show that AdaSent is able to automatically learn the appropriate representation for the task at hand, we visualize the first two principal components (obtained by PCA) of the vector with the largest weight in the hierarchicy for each sentence in the dataset.
Fig.~\ref{fig:pca} shows the projected features from AdaSent (left column) and cBoW (right column) for SUBJ (1st row), MPQA (2nd row) and TREC (3rd row). During training, the model implicitly learns a data representation that enables better prediction. This property of AdaSent is very interesting since we do not explicitly add any separation constraint into our objective function to achieve this.

\section{Conclusion}
\label{sec:conclusion}
In this paper, we propose AdaSent as a new hierarchical sequence modeling approach. AdaSent explores a new direction to represent a sequence by a multi-scale hierarchy instead of a flat, fixed-length, continuous vector representation. The analysis and the empirical results demonstrate the effectiveness and robustness of AdaSent in short sequence modeling. Qualitative results show that AdaSent can learn to represent input sequences depending on the task at hand.

\section*{Acknowledgments}
This work was done when the first and third authors were respectively an intern and a visiting scholar at Noah's Ark Lab, Huawei Technology, Hong Kong. Han Zhao thanks Tao Cai and Baotian Hu at Noah's Ark Lab for their technical support and helpful discussions. This work is supported in part by China National 973 project 2014CB340301.

\bibliographystyle{named}
\bibliography{references}

\end{document}